%% file: main.tex
\documentclass[10pt,twocolumn,letterpaper]{article}

\usepackage{cvpr}              
\usepackage{multirow}
\usepackage[table]{xcolor}
\input{preamble}
\definecolor{cvprblue}{rgb}{0.21,0.49,0.74}
\usepackage[pagebackref,breaklinks,colorlinks,allcolors=cvprblue]{hyperref}

\def\paperID{37027} 
\def\confName{CVPR}
\def\confYear{2026}

\title{Discriminative Perception via Anchored Description for Reasoning Segmentation}

\author{
    Tao Yang\thanks{Equal contribution.},\   
    Qing Zhou\footnotemark[1],\ 
    Yanliang Li,\ 
    Qi Wang\thanks{Corresponding author.}\\
    Northwestern Polytechnical University\\
    {\tt\small \{taoytao, mrazhou, yanliangli\}@mail.nwpu.edu.cn, crabwq@gmail.com}
}

\usepackage{cuted} 
\usepackage{capt-of} 

\begin{document}
\maketitle
\input{sec/0_abstract}    
\input{sec/1_intro}
\input{sec/2_relatedwork}
\input{sec/3_finalcopy}
\input{sec/4_experiment}
\input{sec/5_conclusion}
{
    \small
    \bibliographystyle{ieeenat_fullname}
    \bibliography{main}
}

\input{sec/X_suppl}

\end{document}

%% file: sec/0_abstract.tex
\begin{abstract}
Reasoning segmentation increasingly employs reinforcement learning to generate explanatory reasoning chains that guide Multimodal Large Language Models. While these geometric rewards are primarily confined to guiding the final localization, they are incapable of discriminating whether the reasoning process remains anchored on the referred region or strays into irrelevant context. Lacking this discriminative guidance, the model's reasoning often devolves into unfocused and verbose chains that ultimately fail to disambiguate and perceive the target in complex scenes. This suggests a need to complement the RL objective with Discriminative Perception, an ability to actively distinguish a target from its context. To realize this, we propose DPAD to compel the model to generate a descriptive caption of the referred object, which is then used to explicitly discriminate by contrasting the caption's semantic relevance to the referred object against the wider context. By optimizing for this discriminative capability, the model is forced to focus on the unique attributes of the target, leading to a more converged and efficient reasoning chain. The descriptive caption also serves as an interpretability rationale that aligns with the segmentation. Experiments on the benchmarks confirm the validity of our approach, delivering substantial performance gains, with the cIoU on ReasonSeg increasing by 3.09\% and the reasoning chain length decreasing by approximately 42\%. Code is available at \url{https://github.com/mrazhou/DPAD}.
\end{abstract}

%% file: sec/1_intro.tex
\section{Introduction}
\begin{figure}[ht]
  \centering
  \includegraphics[width=\linewidth]{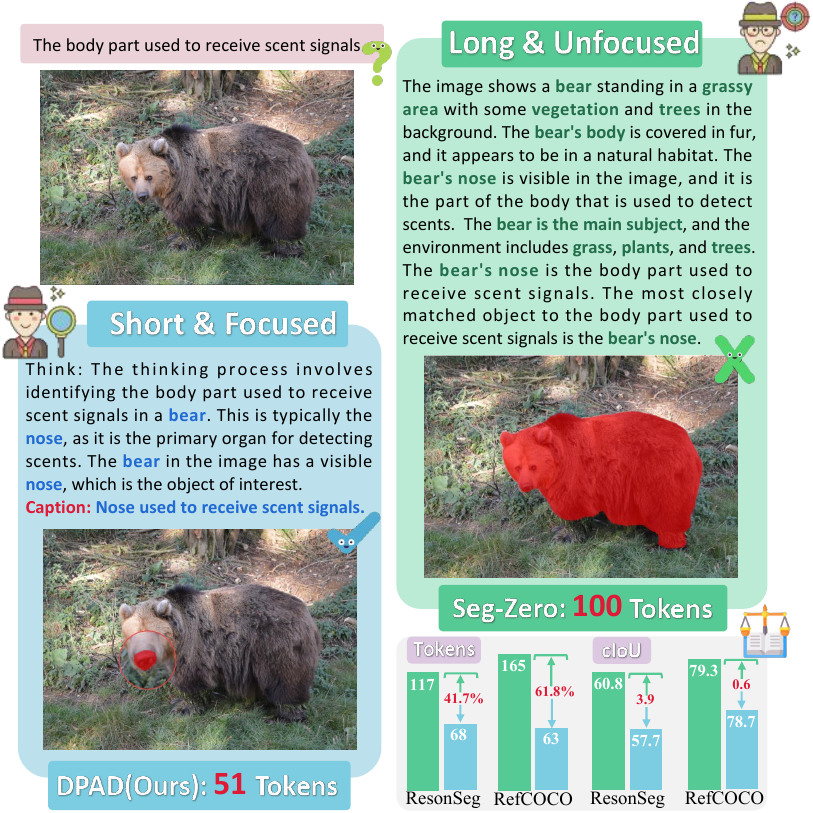}
  \caption{Comparison between Seg-Zero's unfocused reasoning (\textbf{green}), which leads to lower performance, and DPAD's focused reasoning (\textbf{blue}), which is guided by Discriminative Perception to achieve substantial improvements.}
  \label{fig:1}
\end{figure}
Reasoning segmentation (RS) \cite{wang2024llm,wang2024segllm,li2023logicseg} challenges models to generate precise pixel-level masks by interpreting nuanced, context-dependent language queries. The task requires the model to perform inference, deciphering implicit relational and attributive cues within a query to isolate the target within a complex visual scene. The advent of Multimodal Large Language Models (MLLMs) has been pivotal in addressing this challenge \cite{liu2025seg,shen2025reasoning}, leading to architectures that bridge their powerful reasoning capabilities with specialized segmentation modules to achieve language-guided, pixel-level understanding.

Early architectures like the pioneering LISA \cite{lai2024lisa} relied on Supervised Fine-Tuning (SFT). The limited out-of-distribution (OOD) generalization of this approach \cite{li2023logicseg,liu2025seg}, however, catalyzed a paradigm shift towards Reinforcement Learning (RL) to cultivate more generalizable reasoning \cite{huang2024alignsam,chen2024focus,shen2025reasoning}. Within this shift, Group-Relative Policy Optimization (GRPO) \cite{guo2025deepseek} has emerged as a compelling choice, demonstrating substantial advantages for fine-tuning models on task-specific rewards \cite{ting2025semantic,wang2025pixelthink,you2025seg,liu2025visionreasoner}. This RL approach facilitated emergent reasoning, where models autonomously generate Chains-of-Thought, directly improving OOD adaptability in visual tasks. Drawing on this emergent reasoning, several works \cite{liu2025seg,wang2025unified,wang2025pixelthink,shen2025reasoning} in RS have demonstrated that a pure RL strategy can generate reasoning chains to navigate cluttered visual contexts, achieving substantial zero-shot performance improvements on challenging benchmarks such as ReasonSeg. 

Despite their success, these RL-driven paradigms are fundamentally limited by a core learning signal that relies on reward functions based on geometric metrics like IoU and L1 distance \cite{huang2025sam}. While these geometric metrics are primarily confined to guiding the final localization, they offer only indirect or insufficient incentives for the intermediate reasoning process. Indeed, they are incapable of discriminating whether the reasoning process remains anchored on the referred region or strays into irrelevant context. Lacking this discriminative guidance, the model's reasoning can often devolve into chains that are divergent, unanchored, and verbose. This deficiency is particularly detrimental when effective disambiguation and discerning perception are required, as a divergent reasoning chain inevitably incorporates irrelevant contextual distractors, polluting the semantic cues the model relies on \cite{aggarwal2025l1,sui2503stop,qu2025survey}. As a result, the model struggles to isolate the true target from a complex scene, as shown in Figure \ref{fig:1}. To address this critical gap, we propose a complementation in the learning objective for RS itself—moving beyond mere geometric accuracy to actively cultivate the model's ability to distinguish the target from its context, a capability termed Discriminative Perception. 

We realize Discriminative Perception through our method, DPAD (\textbf{D}iscriminative \textbf{P}erception via \textbf{A}nchored \textbf{D}escription). DPAD introduces a discriminative signal to operate in synergy with the geometric rewards by compelling the model to generate an anchored and descriptive caption, subsequently quantifying this caption's discriminative capability through an innovative criterion that contrasts its semantic relevance to the anchored region against its relevance to the wider image context. By optimizing for this discriminative capability, the model's internal reasoning is implicitly rendered more focused and efficient, leading to an average reduction of approximately 42\% in the length of the generated reasoning chains, as shown in Figure \ref{fig:1}. Moreover, the caption offers a transparent explanation for the segmentation, improving interpretability. We conduct extensive experiments on benchmark datasets to validate our approach, demonstrating that it delivers substantial performance gains alongside a more focused and efficient reasoning process.

To summarize as follows:
\begin{itemize}
    \item We introduce Discriminative Perception and propose DPAD, which incentivizes focused reasoning via contrastive discrimination based on an anchored description.
    \item Extensive experiments on challenging reasoning segmentation benchmarks validate that DPAD achieves substantial performance gains.
    \item Moreover, DPAD provides interpretability through a co-generated descriptive caption and reduces the average length of reasoning chains by approximately 42\%.
\end{itemize}

%% file: sec/2_relatedwork.tex
\section{Related Work}
\subsection{Image Segmentation}
Image segmentation \cite{xiong2019upsnet,cheng2022masked}, a fundamental task in computer vision, involves partitioning an image into pixel-level masks. The task has evolved from general semantic segmentation to more nuanced, language-driven objectives. A key development in this area is Referring Expression Segmentation \cite{yang2022lavt,liu2023gres}, which requires segmenting an object specified by a direct natural language phrase and is evaluated on standard benchmarks like RefCOCO, RefCOCO+, and RefCOCOg \cite{yu2016modeling}. Pushing this frontier further, RS \cite{chen2024focus,ren2024pixellm} challenges models to perform pixel-level localization based on more complex language instructions that require inference. This task demands a deep synthesis of visual perception and language comprehension to decipher relational and abstract cues, thus necessitating models with powerful reasoning abilities.

\subsection{Multimodal Large Language Models} The development of MLLMs \cite{li2024mini} represents a significant breakthrough in artificial intelligence, bridging the gap between visual perception and natural language understanding. Pre-trained on vast image-text datasets, these models build internal representations that capture a deep semantic alignment between modalities, allowing them to excel at diverse tasks like visual question answering \cite{huynh2025visual}, image captioning \cite{liu2023visual,Zhousen2024,YANG2026dia}, and complex reasoning \cite{chen2024sam4mllm}. Their ability to comprehend nuanced, context-dependent queries establishes them as a foundational technology for advanced vision-language tasks. These powerful, general-purpose reasoning capabilities, once demonstrated on high-level, holistic image tasks, now pave the way for specialized downstream applications that demand fine-grained, pixel-level understanding, such as reasoning segmentation \cite{ren2024pixellm}.

\subsection{Reasoning Segmentation with MLLMs} Reasoning segmentation leverages the advanced capabilities of MLLMs to perform pixel-level localization based on complex language instructions. The development of this field has undergone a critical paradigm shift from Supervised Fine-Tuning (SFT) to Reinforcement Learning (RL). Early representative methods, such as LISA \cite{lai2024lisa}, applied SFT but faced limitations in generalization when dealing with out-of-distribution (OOD) scenarios. To address this issue, the state-of-the-art has shifted towards RL-based methods, represented by Seg-Zero \cite{liu2025seg}. By employing techniques such as Group-Relative Policy Optimization (GRPO), these models can elicit the emergent generation of chains of thought without explicit reasoning supervision, significantly enhancing zero-shot performance. Concurrently, this paradigm introduces new considerations for reward mechanism design. The geometric reward signals that current methods rely on provide only localization guidance and are unable to discriminate the quality of the reasoning chain. During the learning process, a model might find a pathway that achieves the final localization, but this pathway is not necessarily the most focused or insightful.

%% file: sec/3_finalcopy.tex
\begin{figure*}[t]
  \centering
  \includegraphics[width=\textwidth]{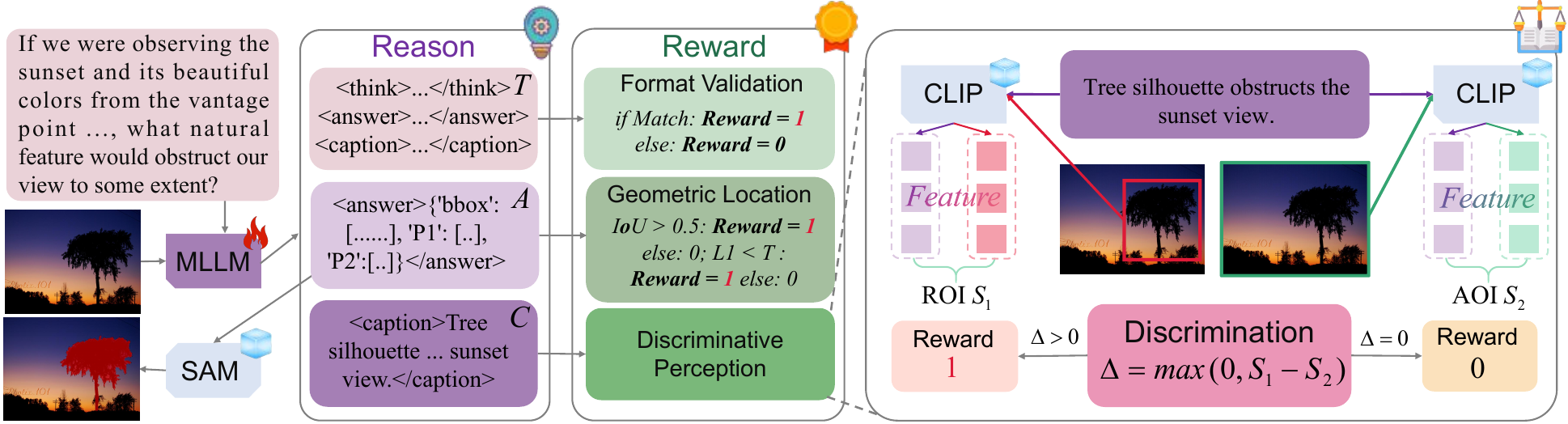}
  \caption{An overview of the DPAD framework, where a MLLM generates a reasoning chain ($T$), a geometric localization ($A$), and an anchored descriptive caption ($C$). The core of our method is the Discriminative Perception reward, which uses the generated caption to derive a Region of Interest (ROI) Score ($S\textsubscript{1}$) and an All of Image (AOI) Score ($S\textsubscript{2}$). The contrast between these scores provides a binary reward signal that incentivizes the model to generate focused reasoning capable of distinguishing the target from its context.}
  \label{fig:2}
\end{figure*}

\section{Methodology}
This section details our proposed method, DPAD. Within an RL-driven framework, DPAD complements rewards for format adherence and geometric accuracy by fostering the model's ``Discriminative Perception'' for more focused and efficient reasoning segmentation.

\subsection{Problem Formulation}

Reasoning segmentation, the task of generating a mask $M$ from a given image $I$ and a language query $Q$, is formulated in our work within an RL framework built upon a decoupled architecture. In this paradigm, a trainable MLLM acts as a reasoning policy, $\pi$, to produce localization prompts. These prompts are then consumed by a frozen segmentation model to render the final mask.

For a given state $s = (I, Q)$, this policy is trained to generate a token sequence $Y$ containing a reasoning chain $T$, a geometric localization $A$, and an anchored descriptive caption $C$, with the generation process expressed as:
\begin{equation}
    Y = (T, A, C) \sim \pi(\cdot|s)
    \label{eq:generation_process}
\end{equation}
The final mask $M$ is then deterministically produced by the localization $A$ via a frozen segmentation module, $\mathcal{F}_{seg}$, such that $M = \mathcal{F}_{seg}(I, A)$. The overarching objective of the RL framework is to learn an optimal policy $\pi^*$ that maximizes the expected final reward, $R_{final}$. This reward function directly evaluates the quality of the generated sequence $Y$. The optimization objective is formally defined as:
\begin{equation}
    \pi^* = \arg\max_{\pi} \mathbb{E}_{Y \sim \pi(\cdot|s)} [R_{final}(Y, s)]
    \label{eq:optimization_objective}
\end{equation}

\subsection{Discriminative Perception}
To address the unfocused and verbose reasoning chains produced by existing RL frameworks, we introduce DPAD (Discriminative Perception via Anchored Description), as shown in Figure~\ref{fig:2}. Our method stimulates the model's ``Discriminative Perception'' by requiring the MLLM policy to generate an \textit{anchored descriptive caption} for its localization, which in turn enables a discriminative reward. This process compels the generation of more focused and efficient reasoning chains.

\textbf{Anchored Descriptive Caption.} In addition to the reasoning chain ($T$) and geometric localization ($A$), our framework requires the MLLM policy to generate a concise and anchored caption ($C$). This caption's purpose is twofold. First, it serves as a human-readable rationale that explains the model's final localization, enhancing the framework's overall interpretability. Second, and more critically, it acts as the semantic foundation for our reward signal, providing the necessary content to evaluate the model's discriminative ability. The caption is considered ``anchored'' because it is prompted to describe the object identified by the model's own geometric localization output ($A$).

\textbf{Discriminative Perception Reward.} This reward is the technical core of DPAD, designed to quantify and incentivize the model's discriminative perception. As illustrated in Figure~\ref{fig:2}, this reward, $R_{\mathrm{dpad}}$, is calculated by contrasting the caption's relevance to the target versus the broader context.
To achieve this, we define the Region of Interest (ROI) as the image crop bounded by the ground-truth box, and the All of Image (AOI) as the entire image. Using a pre-trained vision-language model (e.g., CLIP), we extract semantic features from both the text and the image modalities. Let $V_C$ denote the text feature embedding of the caption $C$, and let $V_{\mathrm{ROI}}$ and $V_{\mathrm{AOI}}$ denote the visual feature embeddings of the ROI and AOI, respectively. The semantic similarity function between a generic text feature vector $V_T$ and a visual feature vector $V_I$ is defined as follows:
\begin{equation}
    \mathrm{Sim}(V_T, V_I) = \frac{V_T \cdot V_I}{\|V_T\| \|V_I\|}
    \label{eq:similarity}
\end{equation}
Based on this, we derive ROI Score, $S_1 = \mathrm{Sim}(V_C, V_{\mathrm{ROI}})$, and AOI Score, $S_2 = \mathrm{Sim}(V_C, V_{\mathrm{AOI}})$. An effective caption must be more relevant to the ROI than to the image as a whole. We thus formulate our final discriminative signal, $\Delta$, to reflect this principle:
\begin{equation}
    \Delta = \max(0, S_1 - S_2)
    \label{eq:delta_signal}
\end{equation}
Here, the term $(S_1 - S_2)$ quantifies the discriminative power of the caption. The discriminative reward, $R_{\mathrm{dpad}}$, is then formalized as follows:
\begin{equation}
    \label{eq:dpad_reward}
    R_{\mathrm{dpad}} =
    \begin{cases}
        1, & \text{if } \Delta > 0 \\
        0, & \text{if } \Delta = 0
    \end{cases}
\end{equation}

\subsection{Final Reward and Optimization}
The complete reward signal, $R_{\mathrm{final}}$, is a composite function designed to guide the policy from multiple perspectives, as illustrated in Figure~\ref{fig:2}. It is composed of three components.

\textbf{Format Validation Reward ($R_{\mathrm{format}}$).} This reward ensures the model's outputs adhere to a predefined structure. We employ three specific checks. First, a regular expression verifies the presence and order of the \texttt{<think>}, \texttt{<answer>}, and \texttt{<caption>} tags. Second, a parser validates the JSON object within the \texttt{<answer>} tag, ensuring it contains the required keys (``bbox'', ``points\_1'', ``points\_2'') and that their values are correctly formatted. Finally, a check confirms the presence of an anchored descriptive caption ($C$) within the model's output. All checks return a binary reward of 1.0 for success and 0.0 for failure, ensuring strict adherence to the desired output format.

\textbf{Geometric Location Reward ($R_{\mathrm{geo}}$).} This reward evaluates the geometric accuracy of the localization output ($A$) to enhance spatial precision. It is a composite of several sub-rewards. An IoU-based reward is calculated, yielding a score of 1.0 if the Intersection over Union between the predicted and ground-truth bounding boxes exceeds 0.5, and 0.0 otherwise. We also employ $L_1$ distance-based rewards for both the bounding box and key points. These provide a reward of 1.0 if the distance is below a set threshold and 0.0 otherwise. This combination of metrics ensures the localization is both spatially accurate and correctly positioned.

The final reward is formulated as follows:
\begin{equation}
    R_{\mathrm{final}} = R_{\mathrm{format}} + R_{\mathrm{geo}} + R_{\mathrm{dpad}}
    \label{eq:final_reward}
\end{equation}

\textbf{Optimization.} The MLLM policy, $\pi$, is then fine-tuned using GRPO to maximize the expected $R_{\mathrm{final}}$. By optimizing this composite objective, the policy is compelled to generate reasoning chains that are not only geometrically accurate but also semantically focused. This process implicitly prunes divergent and verbose thoughts, forcing the model to concentrate on the unique attributes that distinguish the target, thereby leading to more efficient reasoning and improved segmentation performance.

%% file: sec/4_experiment.tex
\begin{table}[t]
\centering
\caption{Performance comparison on existing benchmarks. Symbol $\dagger$ denotes scores reported in the paper while $\ast$ denotes our reproduction with official code and model checkpoint. Our method achieves improvements on ReasonSeg.}
\label{tab:reasonseg}
\begin{tabular}{c|cc|cc}
\toprule
\multirow{2}{*}{Method} & \multicolumn{2}{c|}{val} & \multicolumn{2}{c}{test} \\
                        & gIoU & cIoU & gIoU & cIoU \\
\midrule\midrule
\multicolumn{1}{l|}{OVSeg}                       & 28.5 & 18.6 & 26.1 & 20.8 \\
\multicolumn{1}{l|}{ReLA}                        & 22.4 & 19.9 & 21.3 & 22.0 \\
\multicolumn{1}{l|}{Grounded-SAM}                & 26.0 & 14.5 & 21.3 & 16.4 \\
\multicolumn{1}{l|}{LISA-7B-LLaVA1.5}             & 53.6 & 52.3 & 48.7 & 48.8 \\
\multicolumn{1}{l|}{LISA-13B-LLaVA1.5}            & 57.7 & 60.3 & 53.8 & 50.8 \\
\multicolumn{1}{l|}{SAM4MLLM}                    & 46.7 & 48.1 & -    & -    \\
\multicolumn{1}{l|}{Qwen2.5VL-3B + SAM2}      & 53.8 & 44.1 & 47.6 & 37.4 \\
\multicolumn{1}{l|}{Seg-Zero-7B$^\dagger$}      & 62.6 & 62.0 & 57.5 & 52.0 \\
\midrule
\multicolumn{1}{l|}{Seg-Zero-7B$^*$}                & 60.9 & 57.3 & 57.7 & 54.4 \\
\rowcolor{gray!20}
\multicolumn{1}{l|}{\textbf{DPAD-7B (ours)}} & \textbf{63.1} & \textbf{61.2} & \textbf{60.8} & \textbf{57.5} \\
\bottomrule
\end{tabular}
\end{table}

\begin{table}[t]
\centering
\caption{Performance comparison on existing benchmarks. Symbol $\dagger$ denotes scores reported in the paper while $\ast$ denotes our reproduction with official code and model checkpoint. Our method achieves improvements in cIoU on RefCOCO, RefCOCO+, and RefCOCOg.}
\label{tab:refcoco_results}
\setlength{\tabcolsep}{4pt} 
\begin{tabular}{c|ccc}
\toprule
\multirow{2}{*}{Method} & RefCOCO & RefCOCO+ & RefCOCOg \\
                        & testA      & testA       & test         \\
\midrule\midrule
\multicolumn{1}{l|}{LAVT}                       & 75.8 & 68.4 & 62.1 \\
\multicolumn{1}{l|}{ReLA}                       & 76.5 & 71.0 & 66.0 \\
\multicolumn{1}{l|}{LISA-7B}                    & 76.5 & 67.4 & 68.5 \\
\multicolumn{1}{l|}{PixelLM-7B}                 & 76.5 & 71.7 & 70.5 \\
\multicolumn{1}{l|}{MagNet}                     & 78.3 & 73.6 & 69.3 \\
\multicolumn{1}{l|}{Seg-Zero-7B$^\dagger$}      & 80.3 & 76.2 & 72.6 \\
\midrule
\multicolumn{1}{l|}{Seg-Zero-7B$^*$}                & 78.7 & 74.4 & 71.3 \\
\rowcolor{gray!20}
\multicolumn{1}{l|}{\textbf{DPAD-7B (ours)}} & \textbf{79.3} & \textbf{74.7} & \textbf{72.6} \\
\bottomrule
\end{tabular}
\end{table}

\section{Experiments}
\subsection{Experimental Settings}
We adopt a frozen CLIP model (ViT-B/32) as the semantic scorer to extract both textual and visual features for the computation of $R_{\mathrm{dpad}}$.
To rigorously evaluate our proposed method, DPAD, we utilize a range of standard benchmarks for reasoning and referring expression segmentation.

\textbf{Training Data.} Our model is trained exclusively on a small subset of 3,000 samples from the RefCOCOg dataset \cite{yu2016modeling}. This minimal training set is smaller than those used in prior works, which underscores the superior data and learning efficiency of our method. We do not use any datasets with explicit reasoning chain annotations for training.

\textbf{Evaluation Benchmarks.} The performance of our proposed method, DPAD, is evaluated on several benchmarks. We use ReasonSeg as our primary benchmark to assess the model's generalization capability on complex and nuanced queries. Furthermore, the model's robustness and zero-shot transfer capabilities are validated on the standard test splits of RefCOCO, RefCOCO+, and RefCOCOg.

\subsection{Evaluation Metrics}
We evaluate our model from two perspectives: segmentation performance and discriminative perception capability.

\textbf{Segmentation Performance.} We adopt global Intersection over Union (gIoU) and cumulative Intersection over Union (cIoU) to rigorously assess pixel-level mask prediction. Specifically, gIoU averages the IoU scores across all samples to provide a robust instance-level evaluation that weights small and large targets equally. Conversely, cIoU computes the ratio of total intersection to total union areas across the entire dataset, offering a stable macro-level measurement of pixel precision.

\textbf{Discriminative Perception Capability.} To evaluate the discriminative specificity of the model's outputs towards the target, we introduce a suite of semantic metrics for both the descriptive caption ($C$) and the reasoning chain ($T$). For the caption, we measure its similarity to the ROI ($S_1$) and the AOI ($S_2$), deriving a Semantic Signal-to-Noise Ratio ($\mathrm{SNR} = S_1 / S_2$) to quantify its discriminative power. Similarly, to analyze the internal reasoning process, we compute $TS_1$ and $TS_2$ as the similarities of $T$ to the ROI and AOI, respectively, yielding the Reasoning SNR ($\mathrm{TSNR} = TS_1 / TS_2$). High values for $S_1$ and $TS_1$ indicate strong target relevance, while high $\mathrm{SNR}$ and $\mathrm{TSNR}$ demonstrate precise contextual specificity.

\begin{figure*}[t]
  \centering
  \includegraphics[width=\textwidth]{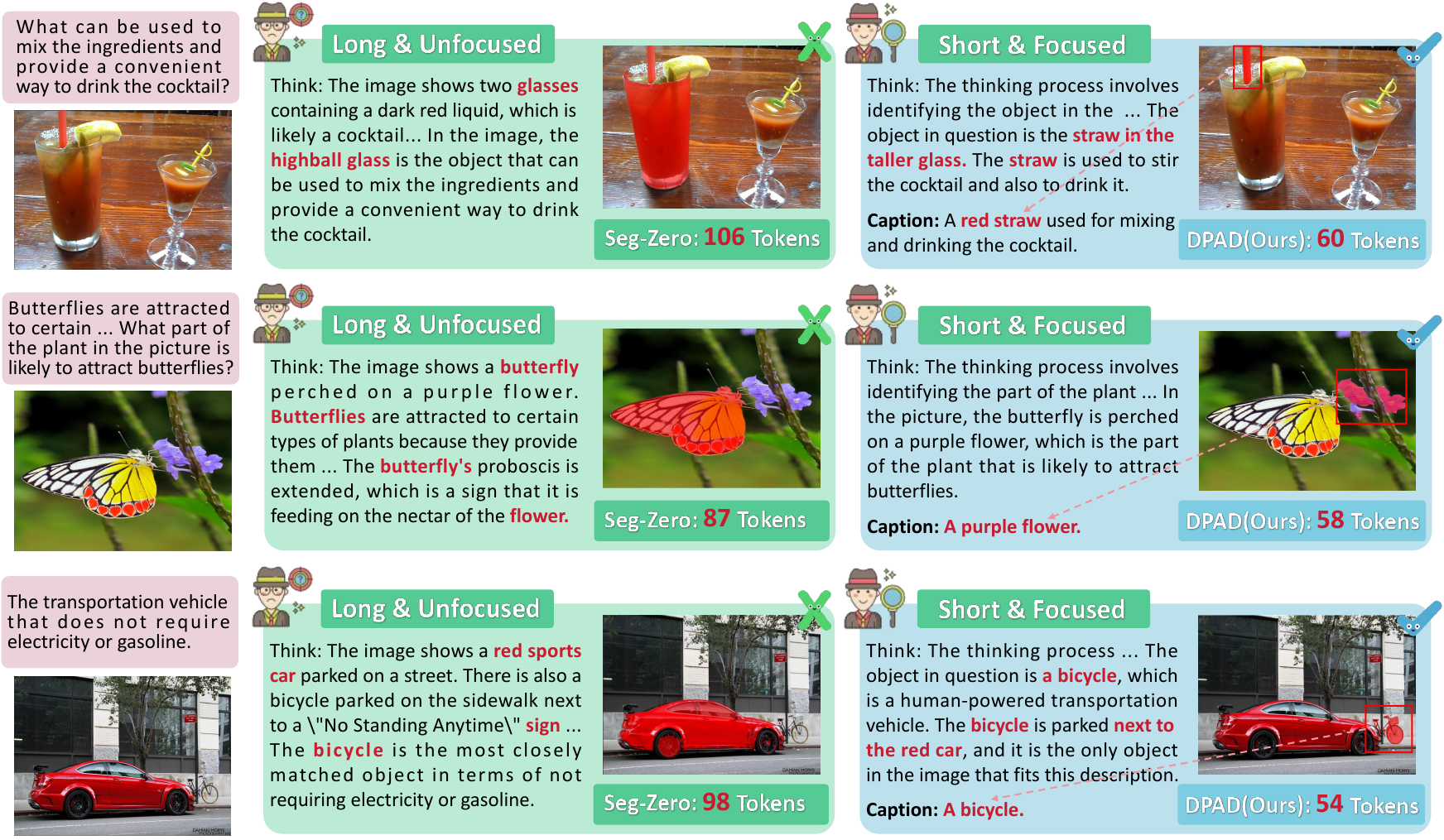}
  \caption{Qualitative comparison between the Seg-Zero and our DPAD. The figure illustrates examples where Seg-Zero produces ``Long \& Unfocused'' reasoning chains that often stray into irrelevant context before identifying the target. In contrast, DPAD generates ``Short \& Focused'' chains that are more converged and efficient. This demonstrates how DPAD's Discriminative Perception leads to more precise reasoning and a significant reduction in token count.}
  \label{fig:example_figure}
\end{figure*}

\begin{table}[htbp]
  \centering
  \caption{Performance and efficiency across different Query Types on ReasonSeg.}
  \label{tab:query_type}
  \begin{tabular}{lcccc}
    \toprule
    \multirow{2}{*}{Query} & \multicolumn{2}{c}{gIoU ($\uparrow$)} & \multicolumn{2}{c}{\#Tok ($\downarrow$)} \\
    \cmidrule(lr){2-3} \cmidrule(lr){4-5}
              & Base  & DPAD           & Base   & DPAD \\
    \midrule
    Attribute & 61.22 & \textbf{64.29} & \phantom{0}97.76  & \textbf{59.95} \\
    Relation  & 58.47 & \textbf{62.17} & \phantom{0}91.73  & \textbf{63.00} \\
    Logic     & 44.92 & \textbf{47.03} & 110.22 & \textbf{68.30} \\
    \bottomrule
  \end{tabular}
\end{table}

\begin{table}[htbp]
  \centering
  \caption{Performance and efficiency across different Difficulty levels on ReasonSeg.}
  \label{tab:difficulty}
  \begin{tabular}{lcccc}
    \toprule
    \multirow{2}{*}{Difficulty} & \multicolumn{2}{c}{gIoU ($\uparrow$)} & \multicolumn{2}{c}{\#Tok ($\downarrow$)} \\
    \cmidrule(lr){2-3} \cmidrule(lr){4-5}
              & Base  & DPAD           & Base   & DPAD \\
    \midrule
    Easy      & 68.37 & \textbf{73.04} & \phantom{0}91.62 & \textbf{59.28} \\
    Medium    & 58.91 & \textbf{61.69} & \phantom{0}96.29 & \textbf{61.66} \\
    Hard      & 45.19 & \textbf{47.48} & 105.06           & \textbf{62.17} \\
    \bottomrule
  \end{tabular}
\end{table}

\begin{table}[t]
\centering
\caption{Comparison of discriminative perception metrics for DPAD and Seg-Zero on ReasonSeg.}
\label{tab:new_metrics_cline_reasonseg}
\setlength{\tabcolsep}{10pt} 
\begin{tabular}{c|cc|cc}
\toprule
\multirow{2}{*}{Metric} & \multicolumn{2}{c|}{Seg-zero-7B} & \multicolumn{2}{c}{\textbf{DPAD-7B (ours)}} \\
                            & val      & test     & val      & test     \\
\midrule\midrule
Tokens\#                    & 117.50   & 111.90   & \textbf{68.49}    & \textbf{68.52}    \\
S\textsubscript{1}          & -        & -        & 25.54    & 25.53    \\
S\textsubscript{2}          & -        & -        & 22.77    & 22.63    \\
SNR                         & -        & -        & \textbf{1.15}     & \textbf{1.16}     \\
TS\textsubscript{1}         & 21.83    & 23.18    & 22.14    & 22.07    \\
TS\textsubscript{2}         & 22.37    & 24.74    & 21.28    & 21.31    \\
TSNR                        & 0.98     & 0.94     & \textbf{1.05}     & \textbf{1.04}     \\
\bottomrule
\end{tabular}
\end{table}

\subsection{Implementation Details}
\textbf{Architecture.} Our framework follows a decoupled design. We use Qwen2.5-VL \cite{bai2025qwen2} as the reasoning model and a frozen SAM2-Large \cite{ravi2024sam} as the segmentation module.

\textbf{Training.} We use a batch size of 16 and a sampling number of 8 for each training step. The initial learning rate is set to $1 \times 10^{-6}$ with a weight decay of 0.01. The reward weights $\lambda$ are fixed at $1.0$ by default. The discriminative reward $R_{\mathrm{dpad}}$ is calculated using a frozen CLIP (ViT-B/32) model, which is entirely discarded during inference.

\subsection{Experimental Results}
\textbf{Segmentation Performance.} We compare our method, DPAD, against a comprehensive suite of state-of-the-art approaches, including OVSeg \cite{liang2023open}, ReLA \cite{liang2023open}, Grounded-SAM \cite{ren2401grounded}, LISA-7B-LLaVA1.5 \cite{lai2024lisa}, LISA-13B-LLaVA1.5 \cite{lai2024lisa}, SAM4MLLM \cite{chen2024sam4mllm}, Qwen2.5VL-3B + SAM2 \cite{wang2025pixelthink}, LAVT \cite{yang2022lavt}, PixelLM-7B \cite{ren2024pixellm}, MagNet \cite{chng2024mask}, and Seg-Zero-7B \cite{liu2025seg}.
As demonstrated in Table~\ref{tab:reasonseg} and Table~\ref{tab:refcoco_results}, our DPAD-7B achieves state-of-the-art performance across multiple challenging benchmarks. On the ReasonSeg dataset, DPAD-7B yields substantial improvements over the strong Seg-Zero-7B$^*$ baseline, boosting gIoU from 57.7 to 60.8 (+3.1) and cIoU from 54.4 to 57.5 (+3.1). This significant gain confirms that explicitly cultivating discriminative perception enhances the model's capacity to disambiguate targets from complex distractors.
Furthermore, we evaluate the zero-shot generalization of DPAD-7B on the RefCOCO, RefCOCO+, and RefCOCOg test splits. As shown in Table~\ref{tab:refcoco_results}, our model outpaces the Seg-Zero-7B$^*$ across all three datasets, achieving 79.3 (+0.6) on RefCOCO, 74.7 (+0.3) on RefCOCO+, and 72.6 (+1.3) on RefCOCOg. These consistent improvements underscore that our discriminative reward mechanism inherently strengthens cross-domain robustness and zero-shot transferability.
\begin{table}[t]
\centering
\caption{Comparison of discriminative perception metrics for DPAD and Seg-Zero on RefCOCO testA (o), RefCOCO testA (+), RefCOCOg test (g).}
\label{tab:new_metrics_detailed}
\setlength{\tabcolsep}{5pt} 
\begin{tabular}{c|ccc|ccc}
\toprule
\multirow{2}{*}{Metric} & \multicolumn{3}{c|}{Seg-zero-7B} & \multicolumn{3}{c}{\textbf{DPAD-7B (ours)}} \\
                            & o        & +        & g        & o        & +        & g        \\
\midrule\midrule
Tokens\#                    & 165.1    & 159.9    & 159.8    & \textbf{63.13} & \textbf{62.88} & \textbf{66.79} \\
S\textsubscript{1}          & -        & -        & -        & 24.30    & 24.64    & 25.91    \\
S\textsubscript{2}          & -        & -        & -        & 20.54    & 19.51    & 22.28    \\
SNR                         & -        & -        & -        & \textbf{1.24}  & \textbf{1.34}  & \textbf{1.21}  \\
TS\textsubscript{1}         & 19.66    & 19.69    & 20.15    & 21.83    & 21.30    & 22.67    \\
TS\textsubscript{2}         & 21.41    & 21.36    & 22.50    & 19.82    & 18.98    & 20.98    \\
TSNR                        & 0.92     & 0.93     & 0.90     & \textbf{1.11}  & \textbf{1.14}  & \textbf{1.09}  \\
\bottomrule
\end{tabular}
\end{table}

\textbf{Stratified Analysis.} To further investigate the robustness of our approach, we stratify the ReasonSeg test set by query type (Attribute, Relation, Logic) and difficulty (Easy, Medium, Hard). Following the complexity assessment protocol in \cite{wang2025pixelthink}, we utilize Qwen2.5-VL-72B to categorize the queries based on scene ambiguity and cognitive complexity. As detailed in Table~\ref{tab:query_type} and Table~\ref{tab:difficulty}, DPAD consistently outperforms the baseline across all data splits. 
When examining query types (Table~\ref{tab:query_type}), the baseline exhibits a clear token explosion on complex Logic queries, consuming over 110 tokens on average. DPAD successfully mitigates this verbosity, compressing the reasoning chain by approximately 38\% while still delivering a $+2.11$ gIoU improvement. Furthermore, DPAD achieves even more pronounced accuracy gains on Attribute ($+3.07$ gIoU) and Relation ($+3.70$ gIoU) queries, demonstrating its broad versatility. 
A similarly compelling trend emerges across varying difficulty levels (Table~\ref{tab:difficulty}). On Easy and Medium subsets, DPAD secures substantial performance boosts of $+4.67$ and $+2.78$ gIoU. Notably, on the most challenging Hard scenarios, DPAD yields a $+2.29$ improvement in gIoU while drastically reducing the reasoning token count by nearly 41\% (from 105.06 to 62.17). 
Crucially, while the baseline's token usage fluctuates wildly with task complexity, DPAD maintains a remarkably stable reasoning length ranging only from 59 to 68 tokens across all stratifications. These fine-grained results robustly validate our core premise that pruning divergent thoughts effectively cultivates a generalizable and efficient reasoning paradigm without compromising the model's capacity for complex logical deduction.

\textbf{Discriminative Perception Capability.} Beyond segmentation accuracy, we analyze the model's internal behavior using our proposed metrics to strictly assess discriminative capability and reasoning efficiency. As detailed in Table~\ref{tab:new_metrics_cline_reasonseg} and Table~\ref{tab:new_metrics_detailed}, DPAD-7B consistently achieves $\mathrm{SNR}$ and $\mathrm{TSNR}$ values exceeding the critical threshold of $1.0$ across all test sets, whereas the Seg-Zero-7B baseline consistently falls below this mark. On the ReasonSeg benchmark, for instance, DPAD-7B records an $\mathrm{SNR}$ of $1.16$ and a $\mathrm{TSNR}$ of $1.04$. Surpassing this $1.0$ threshold signifies that the generated text aligns more closely with the target ROI than with the broader image context. This provides direct quantitative evidence that DPAD acquires true discriminative perception, effectively distinguishing the target from background distractors and mitigating the tendency of prior methods to incorporate irrelevant contextual noise.
Furthermore, this enhanced target focus directly translates to vastly improved reasoning efficiency. Evaluated by the average token count, DPAD-7B generates significantly more concise reasoning chains. On the ReasonSeg test set, the required token count drops from $117.90$ to $68.52$ (a $41.7\%$ reduction). A similar trend emerges across the RefCOCO variants, with the RefCOCOg test set exhibiting a substantial decrease from $159.80$ to $66.79$ (a $58.2\%$ reduction). These compelling results lend robust empirical support to our hypothesis that optimizing for discriminative perception implicitly prunes divergent thoughts, thereby compelling the model toward highly converged and efficient reasoning.

\begin{table}[t]
\centering
\caption{Ablation of reward formulations on ReasonSeg. The best results are highlighted in bold.}
\label{tab:ablation_reward}
\setlength{\tabcolsep}{9pt} 
\begin{tabular}{c|cc|cc}
\toprule
\multirow{2}{*}{Method} & \multicolumn{2}{c|}{val} & \multicolumn{2}{c}{test} \\
                        & gIoU   & cIoU   & gIoU   & cIoU   \\
\midrule\midrule
Baseline                & 60.9   & 57.3   & 57.7   & 54.4   \\
+ Difference            & 61.3   & 58.8   & 57.6   & 54.3   \\
+ Scaled                & 62.6   & 54.0   & 58.4   & 57.5   \\
\textbf{DPAD (Binary)}  & \textbf{63.1} & \textbf{61.2} & \textbf{60.8} & \textbf{57.5} \\
\bottomrule
\end{tabular}
\end{table}

\begin{table}[t]
\centering
\caption{Ablation study of different reward formulations in cIoU on RefCOCO, RefCOCO+, and RefCOCOg test sets, showing cIoU scores. Best results are highlighted in bold.}
\label{tab:ablation_refcoco_ciou}
\setlength{\tabcolsep}{4pt}
\begin{tabular}{c|ccc}
\toprule
\multirow{2}{*}{Method} & RefCOCO & RefCOCO+ & RefCOCOg \\
                        & testA      & testA       & test         \\
\midrule\midrule
Baseline               & 78.7 & 74.4 & 71.3 \\
+ Difference           & 77.2 & 73.2 & 70.8 \\
+ Scaled               & 78.4 & 74.6 & 71.6 \\
\textbf{DPAD (Binary)} & \textbf{79.3} & \textbf{74.7} & \textbf{72.6} \\
\bottomrule
\end{tabular}
\end{table}

\begin{figure}[t]
  \centering
  \includegraphics[width=0.48\textwidth]{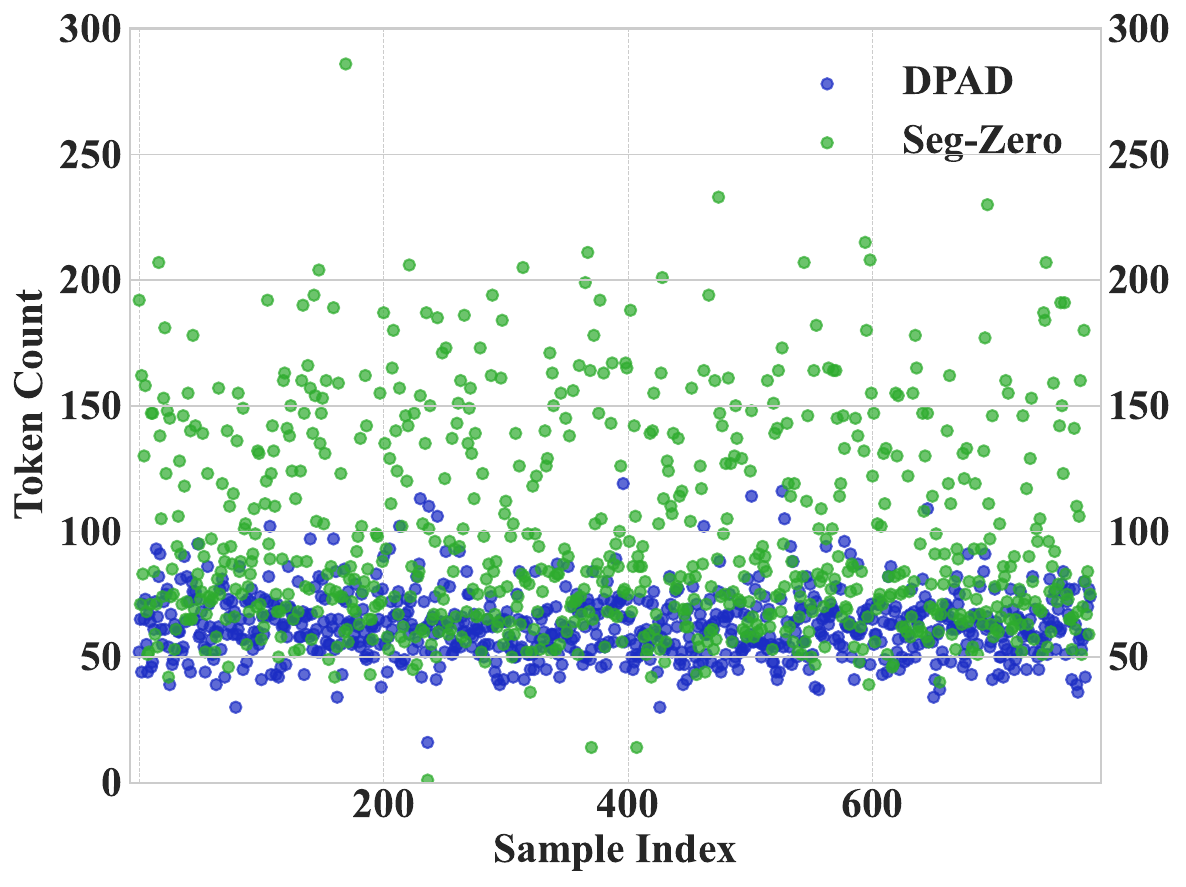}
  \caption{Per-sample token count comparison on the ReasonSeg test. DPAD is shown in blue, and Seg-Zero is shown in green.}
  \label{fig:token_count_scatter}
\end{figure}

\begin{figure}[t]
  \centering
  \includegraphics[width=0.48\textwidth]{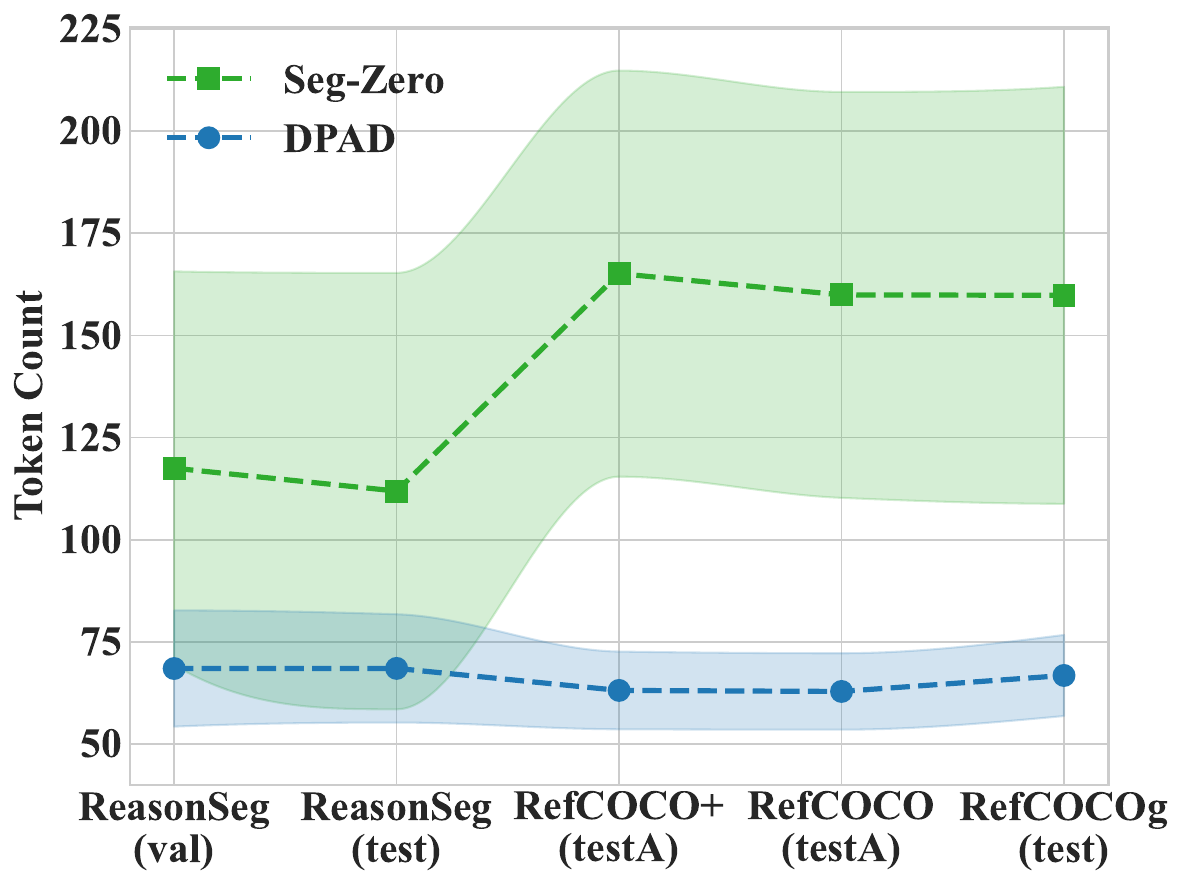}
  \caption{Comparison of the average token counts generated by DPAD and Seg-Zero across five datasets. The lines represent the mean values, while the shaded areas indicate the standard deviation. The results confirm that DPAD accomplishes accurate reasoning using significantly fewer tokens and exhibiting lower variance across all benchmarks.}
  \label{fig:token_comparison}
\end{figure}

\textbf{Qualitative Analysis.} To provide an intuitive understanding of our method's impact, we present a multi-faceted analysis in Figure~\ref{fig:example_figure}, Figure~\ref{fig:token_count_scatter}, and Figure~\ref{fig:token_comparison}.

Figure~\ref{fig:example_figure} presents a direct qualitative comparison between the reasoning processes of DPAD and the Seg-Zero baseline. The visualizations illustrate how DPAD's core principle, Discriminative Perception, addresses the verbose and unfocused chains typical of prior methods. By explicitly incentivizing target-relevant descriptions, our framework actively prunes divergent global scenes, irrelevant objects, and redundant self-confirmations from the internal thought process. For instance, in identifying an object for a cocktail, Seg-Zero strays into irrelevant context, while DPAD immediately identifies the correct object (the straw), effectively cutting the token count from 106 to 60. This powerful disambiguation capability is equally evident in the non-electric vehicle example; DPAD avoids distraction from the visually dominant sports car and correctly locates the bicycle, reducing the token count from 98 to 54.

This efficiency is not an anomaly. Figure~\ref{fig:token_count_scatter} extends this analysis to every sample in the ReasonSeg test set, plotting the token count for each individual inference. The scatter plot reveals that DPAD's token counts (blue) are consistently clustered in a low, tight band, demonstrating stable and highly efficient performance. In contrast, Seg-Zero's counts (green) are not only significantly higher on average but also exhibit much greater variance across samples.

Furthermore, Figure~\ref{fig:token_comparison} provides compelling evidence for how our method's focused reasoning translates into superior efficiency and stability across five distinct benchmarks. As a direct result of our discriminative optimization, DPAD maintains a consistently low average token count (the blue line, approximately 65 tokens) with a narrow variance band, signifying robust, low-variance reasoning. Conversely, Seg-Zero's unfocused process leads to a substantially higher and more volatile token count. This instability is visually evident in its fluctuating average and wide variance band, revealing a high-variance reasoning process that is overly sensitive to dataset-specific complexities.

Overall, these visualizations robustly support our central thesis that by compelling the model to generate an anchored description and optimizing for a discriminative reward, DPAD successfully cultivates a highly focused and efficient reasoning paradigm. This not only improves quantitative performance but also produces transparent reasoning chains devoid of irrelevant contextual noise.

\subsection{Ablation Study}
To investigate the impact of different reward formulations, we conduct an ablation study with results detailed in Table~\ref{tab:ablation_reward} and Table~\ref{tab:ablation_refcoco_ciou}. We compare our final DPAD model equipped with a binary reward against a Baseline trained without any discriminative signal, alongside two alternative continuous reward designs. The first alternative is a ``Difference'' reward formulated as $R = S_1 - S_2$, and the second is a ``Scaled'' reward calculated as $R = S_1 \times \max(0, S_1 - S_2)$.

A clear performance hierarchy emerges across all benchmarks. On the ReasonSeg dataset (Table~\ref{tab:ablation_reward}), all three reward formulations yield improvements over the Baseline. The ``Difference'' reward improves the validation gIoU to 61.3 (+0.4), while the ``Scaled'' variant reaches 62.6 (+1.7). Crucially, our binary reward establishes the best performance among all variants, boosting validation gIoU to 63.1 (+2.2) and cIoU to 61.2 (+3.9). This trajectory extends to the RefCOCO experiments (Table~\ref{tab:ablation_refcoco_ciou}). All variants surpass the Baseline, with the binary reward uniformly securing the highest scores across the three test splits, highlighted by a notable cIoU of 72.6 on the challenging RefCOCOg benchmark.

We attribute the clear superiority of the binary reward to its signal format, which is highly compatible with the internal optimization mechanics of our framework. Specifically, our system---including the Format Validation reward ($R_{\mathrm{format}}$) and the Geometric Location reward ($R_{\mathrm{geo}}$)---fundamentally relies on discrete success-or-failure signals to guide the GRPO-based policy updates. Our binary discriminative reward aligns with this effective paradigm, providing unambiguous feedback on whether the model distinguishes the target. In contrast, while continuous rewards offer more granular semantic information, their inherent numerical noise and scale fluctuations actively interfere with the stable convergence of the RL policy.

%% file: sec/5_conclusion.tex
\section{Conclusion}
In this work, we introduced DPAD, a framework that mitigates unfocused reasoning in segmentation models by fostering discriminative perception. By compelling the MLLM to generate an anchored descriptive caption, DPAD learns to explicitly distinguish targets from their context, yielding a highly converged and efficient reasoning process. Extensive experiments demonstrate that DPAD achieves state-of-the-art performance across multiple challenging benchmarks, including ReasonSeg and the RefCOCO suite. Crucially, these accuracy gains are accompanied by an approximate $42\%$ reduction in reasoning chain length and discriminative signals ($\mathrm{SNR}$ and $\mathrm{TSNR}$) consistently exceeding the critical $1.0$ threshold. Beyond quantitative improvements, DPAD significantly enhances interpretability through its co-generated captions. Ultimately, our findings suggest that optimizing for discriminative perception offers a highly promising pathway toward building accurate, focused, and efficient Multimodal Large Language Models for complex visual reasoning tasks.

\section*{Acknowledgments}
This work was supported by the National Natural Science Foundation of China under Grants 62471394 and U21B2041.

%% file: sec/X_suppl.tex
\clearpage
\setcounter{page}{1}
\maketitlesupplementary

\begin{strip}
    \centering
    \includegraphics[width=0.99\linewidth]{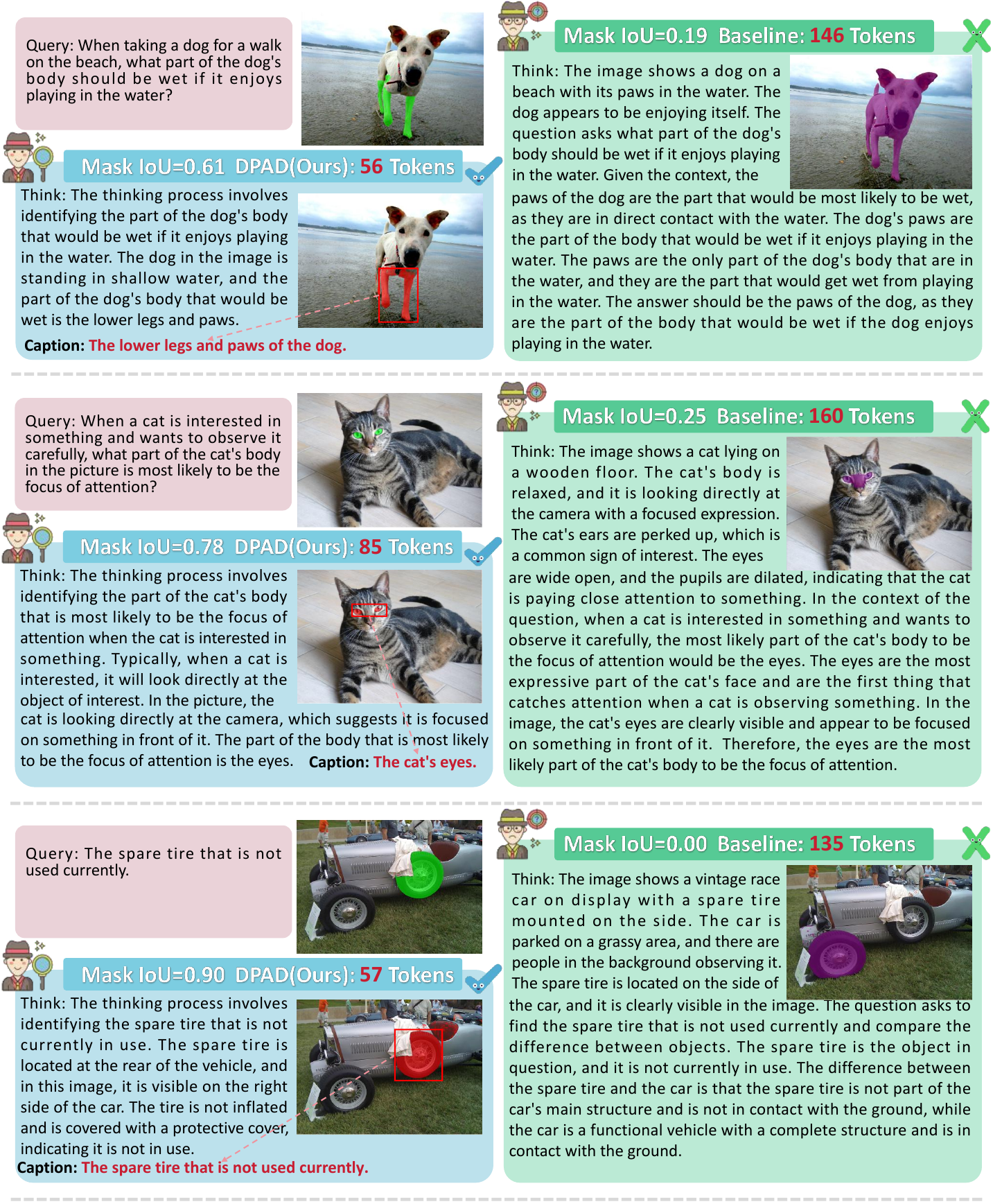}
    \vspace{1em} 
    \includegraphics[width=0.99\linewidth]{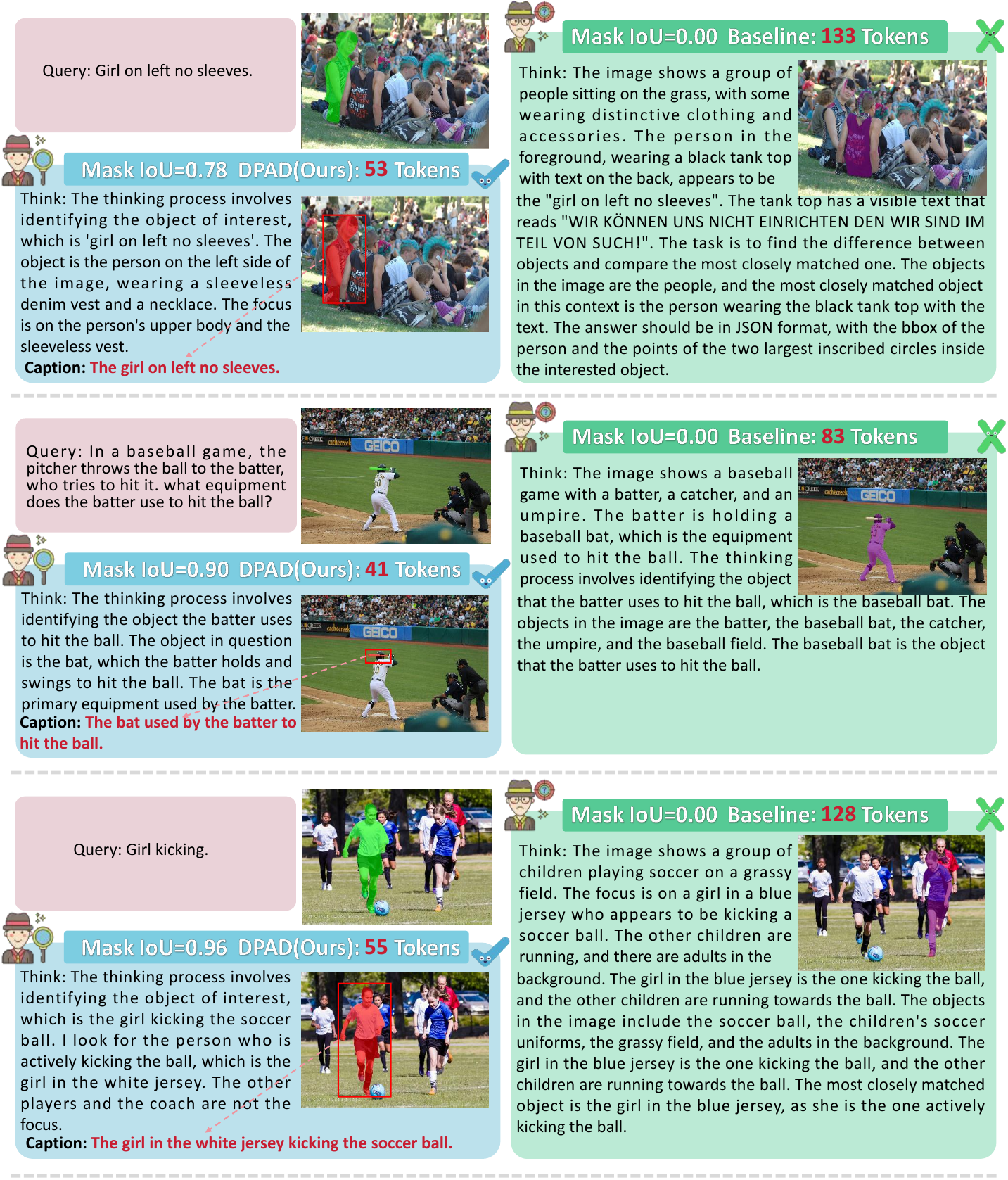}
    
    \captionof{figure}{Additional qualitative comparison. Masks: GT (green), DPAD (red), and baseline (purple).}
    \label{fig:qua_dpad_analysis}
\end{strip}

\begin{figure*}[h!]
    \centering
    \includegraphics[width=0.99\linewidth]{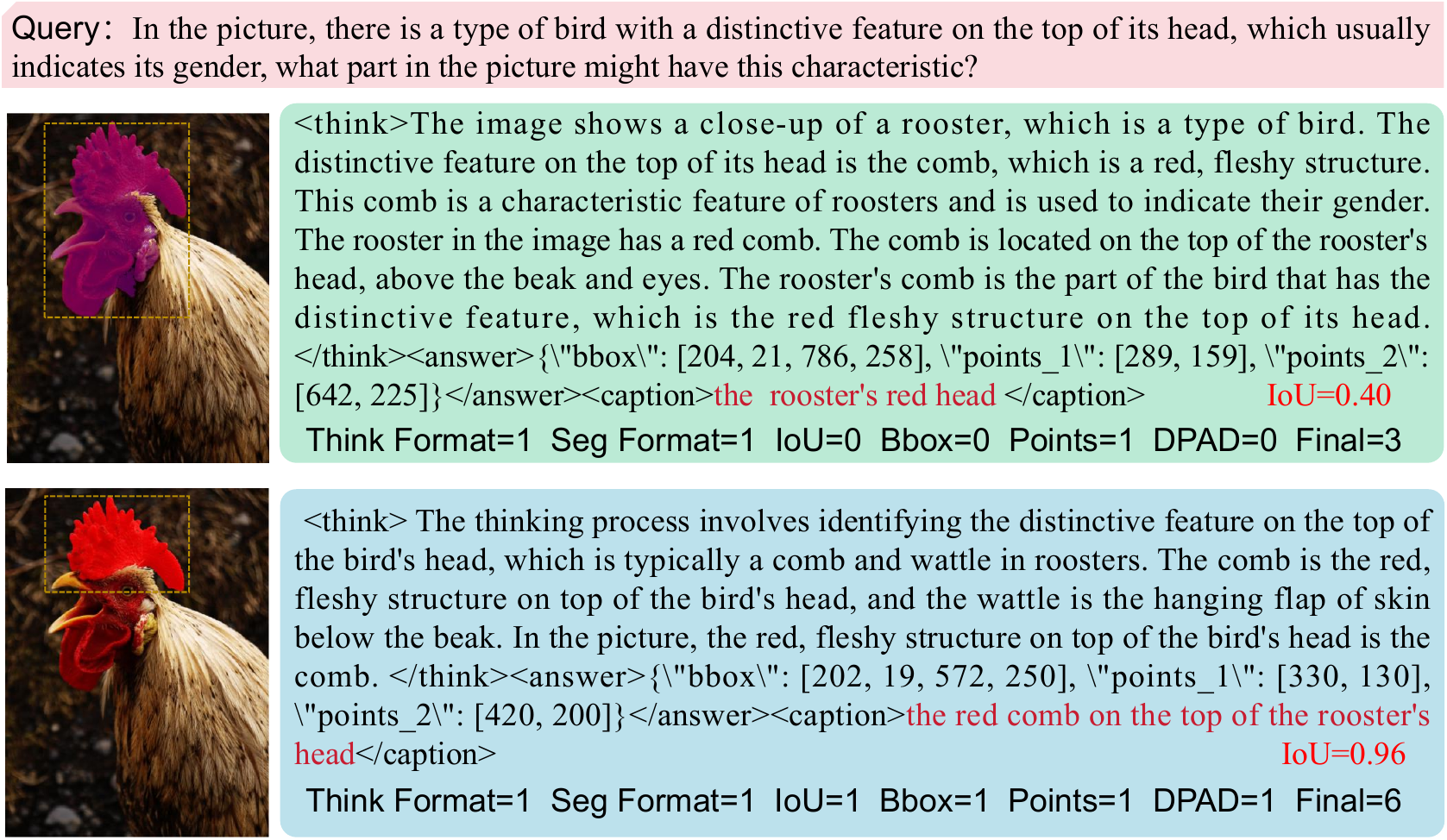}
    \caption{Qualitative demonstration of spatial-semantic alignment evolution via discriminative perception.}
    \label{fig:rooster_evolution}
\end{figure*}

\section{Additional Qualitative Analysis}
\label{sec:qua_dpad_analysis}

It can be observed that models lacking the ``Discriminative Perception'' reward often generate verbose and divergent reasoning chains, as detailed in Figure \ref{fig:qua_dpad_analysis}. Specifically, DPAD significantly reduces reasoning redundancy across the following three core dimensions:

\textbf{Pruning Divergent Global Scenes.} In the baseline method, the model frequently allocates excessive attention to global background descriptions irrelevant to the query. For instance, in the ``spare tire'' and ``girl kicking'' examples, the baseline model consumes a large number of tokens describing the ``vintage race car... parked on a grassy area,'' ``people in the background,'' or ``adults in the background''. In contrast, DPAD, through the contrastive reward mechanism of anchored descriptions, effectively guides the model to disregard these divergent global details and anchor its reasoning directly on the target object (e.g., clearly locating the uninflated spare tire on the side of the car). This achieves accurate segmentation using significantly fewer tokens (e.g., reducing the count from 135 to 57).
    
\textbf{Filtering Irrelevant Objects and Over-identification.} When multiple objects or complex semantics are present in a scene, the baseline model is highly prone to enumerative behavior or deviating from the core task. In the ``baseball game'' example, the baseline not only describes the batter but also redundantly lists the catcher, the umpire, and the baseball field. This is particularly evident in the ``girl on left no sleeves'' example, where the baseline diverges from the visual grounding objective to recognize and transcribe a large block of German text (``WIR KÖNNEN UNS NICHT EINRICHTEN...'') on the person's clothing. DPAD effectively filters out these distractors and precisely converges on the ``bat'' and the ``sleeveless denim vest,'' which not only reduces the token count by more than half but also substantially improves the mask IoU from 0.00 to high-precision levels (0.90 and 0.78, respectively).
    
\textbf{Reducing Over-reasoning and Self-confirmations.} In the examples of the ``dog's paws'' and ``cat's eyes,'' the baseline exhibits a pronounced tendency toward ``self-confirmation.'' The model repeatedly restates the problem context and incorporates superfluous trivial details (such as repeatedly emphasizing that the cat's ``ears are perked up'' and ``pupils are dilated,'' or that the dog ``appears to be enjoying itself''). DPAD, on the other hand, presents a highly ``Short \& Focused'' reasoning paradigm, addressing the core attributes of the target directly and concisely outputting the ``lower legs and paws'' or the ``cat's eyes.''

These additional visualization results provide compelling evidence that, by incentivizing target-relevant descriptions, DPAD structurally prunes divergent scenes, irrelevant objects, and uninformative self-confirmations. This not only substantially compresses the length of the reasoning chains (mitigating the redundancy issues that degrade reasoning efficiency) but, more importantly, provides highly precise localization cues for downstream mask generation by eliminating contextual noise.

\section{Evolution of DPAD Alignment}
\label{sec:supp_dpad_evolution}

In the evolution analysis of the convergence process, we observe how the DPAD reward mechanism calibrates spatial outputs through differentiated feedback on intermediate states (as shown in Figure \ref{fig:rooster_evolution}). During the under-converged stages of training, the model frequently exhibits a disconnect between semantics and geometry. At this stage, if the model relies solely on visual priors to yield a coarse bounding box covering most of the rooster's head, but generates an over-generalized or inaccurate description (e.g., ``the rooster's red head''), our mechanism assigns a reward of $R=0$ to suppress such blind localization lacking rigorous logical support.
When the model demonstrates correct anatomical discrimination during its reasoning process (e.g., explicitly distinguishing between the ``comb on top of the head'' and the ``wattle under the beak'') and generates the precise description ``the red comb on the top of the rooster's head'', the mechanism provides positive feedback of $R=1$. Continuously driven by this mechanism, the model is able to correct the initially ambiguous boundaries, precisely contracting and locking the bounding box onto the comb region, raising the IoU to $0.96$.

%

%% file: main.bib
@String(AAAI = {AAAI})

@inproceedings{lai2024lisa,
  title     = "{Lisa: Reasoning Segmentation via Large Language Model}",
  author    = "Lai, Xin and Tian, Zhuotao and Chen, Yukang and Li, Yanwei and Yuan, Yuhui and Liu, Shu and Jia, Jiaya",
  booktitle = "Proceedings of the IEEE/CVF Conference on Computer Vision and Pattern Recognition",
  pages     = "9579--9589",
  year      = 2024,
}

@article{liu2025seg,
  title   = "{Seg-zero: Reasoning-chain Guided Segmentation via Cognitive Reinforcement}",
  author  = "Liu, Yuqi and Peng, Bohao and Zhong, Zhisheng and Yue, Zihao and Lu, Fanbin and Yu, Bei and Jia, Jiaya",
  journal = "arXiv preprint arXiv:2503.06520",
  year    = 2025,
}

@article{wang2025pixelthink,
  title   = "{PixelThink: Towards Efficient Chain-of-Pixel Reasoning}",
  author  = "Wang, Song and others",
  journal = "arXiv preprint arXiv:2505.23727",
  year    = 2025,
}

@article{shen2025reasoning,
  title   = "{Reasoning Segmentation for Images and Videos: A Survey}",
  author  = "Shen, Yiqing and Li, Chenjia and Xiong, Fei and Jeong, Jeong-O and Wang, Tianpeng and Latman, Michael and Unberath, Mathias",
  journal = "arXiv preprint arXiv:2505.18816",
  year    = 2025,
}

@inproceedings{li2023logicseg,
  title     = "{LogicSeg: Parsing Visual Semantics with Neural Logic Learning and Reasoning}",
  author    = "Li, Liulei and Wang, Wenguan and Yang, Yi",
  booktitle = "Proceedings of the IEEE/CVF International Conference on Computer Vision",
  pages     = "4122--4133",
  year      = 2023,
}

@inproceedings{huang2024alignsam,
  title     = "{AlignSAM: Aligning Segment Anything Model to Open Context via Reinforcement Learning}",
  author    = "Huang, Duojun and Xiong, Xinyu and Ma, Jie and Li, Jichang and Jie, Zequn and Ma, Lin and Li, Guanbin",
  booktitle = "Proceedings of the IEEE/CVF Conference on Computer Vision and Pattern Recognition",
  pages     = "3205--3215",
  year      = 2024,
}

@inproceedings{chen2024focus,
  title     = "{Focus-then-Decide: Segmentation-assisted Reinforcement Learning}",
  author    = "Chen, Chao and Xu, Jiacheng and Liao, Weijian and Ding, Hao and Zhang, Zongzhang and Yu, Yang and Zhao, Rui",
  booktitle = "Proceedings of the AAAI Conference on Artificial Intelligence",
  volume    = "38",
  number    = "10",
  pages     = "11240--11248",
  year      = 2024,
}

@article{you2025seg,
  title={Seg-R1: Segmentation Can Be Surprisingly Simple with Reinforcement Learning},
  author={You, Zuyao and Wu, Zuxuan},
  journal={arXiv preprint arXiv:2506.22624},
  year={2025}
}

@inproceedings{wang2024llm,
  title     = "{LLM-Seg: Bridging Image Segmentation and Large Language Model Reasoning}",
  author    = "Wang, Junchi and Ke, Lei",
  booktitle = "Proceedings of the IEEE/CVF Conference on Computer Vision and Pattern Recognition",
  pages     = "1765--1774",
  year      = 2024,
}

@article{wang2024segllm,
  title   = "{SegLLM: Multi-round Reasoning Segmentation}",
  author  = "Wang, XuDong and Zhang, Shaolun and Li, Shufan and Kallidromitis, Konstantinos and Li, Kehan and Kato, Yusuke and Kozuka, Kazuki and Darrell, Trevor",
  journal = "arXiv preprint arXiv:2410.18923",
  year    = 2024,
}

@article{liu2025visionreasoner,
  title   = "{VisionReasoner: Unified Visual Perception and Reasoning via Reinforcement Learning}",
  author  = "Liu, Yuqi and Qu, Tianyuan and Zhong, Zhisheng and Peng, Bohao and Liu, Shu and Yu, Bei and Jia, Jiaya",
  journal = "arXiv preprint arXiv:2505.12081",
  year    = 2025,
}

@article{wang2025unified,
  title   = "{Unified Multimodal Chain-of-Thought Reward Model through Reinforcement Fine-tuning}",
  author  = "Wang, Yibin and Li, Zhimin and Zang, Yuhang and Wang, Chunyu and Lu, Qinglin and Jin, Cheng and Wang, Jiaqi",
  journal = "arXiv preprint arXiv:2505.03318",
  year    = 2025,
}

@article{ting2025semantic,
  title   = "{Semantic Segmentation with Reward}",
  author  = "Ting, Xie and Huang, Ye and Liu, Zhilin and Duan, Lixin",
  journal = "arXiv preprint arXiv:2505.17905",
  year    = 2025,
}

@article{guo2025deepseek,
  title   = "{Deepseek-RL: Incentivizing Reasoning Capability in LLMs via Reinforcement Learning}",
  author  = "Guo, Daya and Yang, Dejian and Zhang, Haowei and Song, Junxiao and Zhang, Ruoyu and Xu, Runxin and Zhu, Qihao and Ma, Shirong and Wang, Peiyi and Bi, Xiao and others",
  journal = "arXiv preprint arXiv:2501.12948",
  year    = 2025,
}

@article{huang2025sam,
  title   = "{SAM-RL: Leveraging SAM for Reward Feedback in Multimodal Segmentation via Reinforcement Learning}",
  author  = "Huang, Jiaqi and Xu, Zunnan and Zhou, Jun and Liu, Ting and Xiao, Yicheng and Ou, Mingwen and Ji, Bowen and Li, Xiu and Yuan, Kehong",
  journal = "arXiv preprint arXiv:2505.22596",
  year    = 2025,
}

@article{liu2023visual,
  title   = "{Visual Instruction Tuning}",
  author  = "Liu, Haotian and Li, Chunyuan and Wu, Qingyang and Lee, Yong Jae",
  journal = "Advances in Neural Information Processing Systems",
  volume  = "36",
  pages   = "34892--34916",
  year    = 2023,
}

@article{aggarwal2025l1,
  title   = "{L1: Controlling How Long a Reasoning Model Thinks with Reinforcement Learning}",
  author  = "Aggarwal, Pranjal and Welleck, Sean",
  journal = "arXiv preprint arXiv:2503.04697",
  year    = "2025",
}

@article{sui2503stop,
  title   = "{Stop Overthinking: A Survey on Efficient Reasoning for Large Language Models}",
  author  = "Sui, Yang and Chuang, Yu-Neng and Wang, Guanchu and Zhang, Jiamu and Zhang, Tianyi and Yuan, Jiayi and Liu, Hongyi and Wen, Andrew and Zhong, Shaochen and Chen, Hanjie and others",
  journal = "arXiv preprint arXiv:2503.16419",
  year    = 2025,
}

@article{qu2025survey,
  title   = "{A Survey of Efficient Reasoning for Large Reasoning Models: Language, Multimodality, and Beyond}",
  author  = "Qu, Xiaoye and Li, Yafu and Su, Zhaochen and Sun, Weigao and Yan, Jianhao and Liu, Dongrui and Cui, Ganqu and Liu, Daizong and Liang, Shuxian and He, Junxian and others",
  journal = "arXiv preprint arXiv:2503.21614",
  year    = 2025,
}

@inproceedings{xiong2019upsnet,
  title     = "{UPSNet: A Unified Panoptic Segmentation Network}",
  author    = "Xiong, Yuwen and Liao, Renjie and Zhao, Hengshuang and Hu, Rui and Bai, Min and Yumer, Ersin and Urtasun, Raquel",
  booktitle = "Proceedings of the IEEE/CVF Conference on Computer Vision and Pattern Recognition",
  pages     = "8818--8826",
  year      = 2019,
}

@inproceedings{cheng2022masked,
  title     = "{Masked-attention Mask Transformer for Universal Image Segmentation}",
  author    = "Cheng, Bowen and Misra, Ishan and Schwing, Alexander G and Kirillov, Alexander and Girdhar, Rohit",
  booktitle = "Proceedings of the IEEE/CVF Conference on Computer Vision and Pattern Recognition",
  pages     = "1290--1299",
  year      = 2022,
}

@inproceedings{yang2022lavt,
  title     = "{LAVT: Language-aware Vision Transformer for Referring Image Segmentation}",
  author    = "Yang, Zhao and Wang, Jiaqi and Tang, Yansong and Chen, Kai and Zhao, Hengshuang and Torr, Philip HS",
  booktitle = "Proceedings of the IEEE/CVF Conference on Computer Vision and Pattern Recognition",
  pages     = "18155--18165",
  year      = 2022,
}

@inproceedings{yu2016modeling,
  title        = "{Modeling Context in Referring Expressions}",
  author       = "Yu, Licheng and Poirson, Patrick and Yang, Shan and Berg, Alexander C and Berg, Tamara L",
  booktitle    = "European Conference on Computer Vision",
  pages        = "69--85",
  year         = 2016,
  organization = "Springer",
}

@article{li2024mini,
  title   = "{Mini-Gemini: Mining the Potential of Multi-modality Vision Language Models}",
  author  = "Li, Yanwei and Zhang, Yuechen and Wang, Chengyao and Zhong, Zhisheng and Chen, Yixin and Chu, Ruihang and Liu, Shaoteng and Jia, Jiaya",
  journal = "arXiv preprint arXiv:2403.18814",
  year    = 2024,
}

@inproceedings{chen2024sam4mllm,
  title        = "{SAM4MLLM: Enhance Multi-modal Large Language Model for Referring Expression Segmentation}",
  author       = "Chen, Yi-Chia and Li, Wei-Hua and Sun, Cheng and Wang, Yu-Chiang Frank and Chen, Chu-Song",
  booktitle    = "European Conference on Computer Vision",
  pages        = "323--340",
  year         = 2024,
  organization = "Springer",
}

@article{huynh2025visual,
  title   = "{Visual Question Answering: From Early Developments to Recent Advances--A Survey}",
  author  = "Huynh, Ngoc Dung and Bouadjenek, Mohamed Reda and Aryal, Sunil and Razzak, Imran and Hacid, Hakim",
  journal = "arXiv preprint arXiv:2501.03939",
  year    = 2025,
}

@inproceedings{ren2024pixellm,
  title     = "{PixelLM: Pixel Reasoning with Large Multimodal Model}",
  author    = "Ren, Zhongwei and Huang, Zhicheng and Wei, Yunchao and Zhao, Yao and Fu, Dongmei and Feng, Jiashi and Jin, Xiaojie",
  booktitle = "Proceedings of the IEEE/CVF Conference on Computer Vision and Pattern Recognition",
  pages     = "26374--26383",
  year      = 2024,
}

@article{bai2025qwen2,
  title={Qwen2. 5-vl technical report},
  author={Bai, Shuai and Chen, Keqin and Liu, Xuejing and Wang, Jialin and Ge, Wenbin and Song, Sibo and Dang, Kai and Wang, Peng and Wang, Shijie and Tang, Jun and others},
  journal={arXiv preprint arXiv:2502.13923},
  year={2025}
}

@article{ravi2024sam,
  title={Sam 2: Segment anything in images and videos},
  author={Ravi, Nikhila and Gabeur, Valentin and Hu, Yuan-Ting and Hu, Ronghang and Ryali, Chaitanya and Ma, Tengyu and Khedr, Haitham and R{\"a}dle, Roman and Rolland, Chloe and Gustafson, Laura and others},
  journal={arXiv preprint arXiv:2408.00714},
  year={2024}
}

@inproceedings{liang2023open,
  title={Open-vocabulary semantic segmentation with mask-adapted clip},
  author={Liang, Feng and Wu, Bichen and Dai, Xiaoliang and Li, Kunpeng and Zhao, Yinan and Zhang, Hang and Zhang, Peizhao and Vajda, Peter and Marculescu, Diana},
  booktitle={Proceedings of the IEEE/CVF conference on computer vision and pattern recognition},
  pages={7061--7070},
  year={2023}
}

@inproceedings{liu2023gres,
  title={Gres: Generalized referring expression segmentation},
  author={Liu, Chang and Ding, Henghui and Jiang, Xudong},
  booktitle={Proceedings of the IEEE/CVF conference on computer vision and pattern recognition},
  pages={23592--23601},
  year={2023}
}

@article{ren2401grounded,
  title   = "{Grounded SAM: Assembling Open-World Models for Diverse Visual Tasks}",
  author  = "Ren, Tianhe and Liu, Shilong and Zeng, Ailing and Lin, Jing and Li, Kunchang and Cao, He and Chen, Jiayu and Huang, Xinyu and Chen, Yukang and Yan, Feng and others",
  journal = "arXiv preprint arXiv:2401.14159",
  year    = 2024,
}

@inproceedings{chng2024mask,
  title={Mask grounding for referring image segmentation},
  author={Chng, Yong Xien and Zheng, Henry and Han, Yizeng and Qiu, Xuchong and Huang, Gao},
  booktitle={Proceedings of the IEEE/CVF Conference on Computer Vision and Pattern Recognition},
  pages={26573--26583},
  year={2024}
}

@article{Zhousen2024,
  author={Zhou, Qing and Gao, Junyu and Yuan, Yuan and Wang, Qi},
  journal={IEEE Transactions on Geoscience and Remote Sensing}, 
  title={Single-Stream Extractor Network With Contrastive Pre-Training for Remote-Sensing Change Captioning}, 
  year={2024},
  volume={62},
  number={},
  pages={1-14},
}

@article{YANG2026dia,
  author={Yang, Tao and Zhou, Qing and Wang, Qi},
  journal={Pattern Recognition}, 
  title={DIA: Deriving linguistic information from auxiliary languages for remote sensing image captioning}, 
  year={2026},
  volume={171},
  number={},
  pages={112209},
}
